%% file: main.tex
\documentclass[10pt,twocolumn,letterpaper]{article}

\usepackage{cvpr}              

\definecolor{cvprblue}{rgb}{0.21,0.49,0.74}
\usepackage[pagebackref,breaklinks,colorlinks,allcolors=cvprblue]{hyperref}
\usepackage{natbib}
\usepackage{colortbl}
\definecolor{lightgreen}{rgb}{0.84, 1.0, 0.84}
\usepackage{multirow}

\usepackage{pifont} 
\usepackage{booktabs} 
\newcommand{\cmark}{\ding{51}} 
\newcommand{\xmark}{\ding{55}} 
\usepackage{adjustbox}
\usepackage{arydshln}
\usepackage{tabularx}

\def\paperID{15295} 
\def\confName{CVPR}
\def\confYear{2025}

\title{E2LVLM: Evidence-Enhanced Large Vision-Language Model for Multimodal Out-of-Context Misinformation Detection}

\author{
Junjie Wu$^1$ \ \  \  Yumeng Fu$^3$ \ \  \    Nan Yu$^1$  \ \  \   Guohong Fu$^{1,2}$\thanks{Corresponding author.}\\
$^1$School of Computer Science and Technology, Soochow University\\
$^2$Institute of Artificial Intelligence, Soochow University\\
$^3$School of Computer Science and Technology, Harbin Institute of Technology\\
{\tt\small \{20224027010, nyu\}@stu.suda.edu.cn, 24b303004@stu.hit.edu.cn, ghfu@suda.edu.cn}
}

\begin{document}
\maketitle

\input{0_abstract}    
\input{1_intro}
\input{2_related}
\input{3_methodology}
\input{4_experiment}
\input{5_conclusion}
{
    \small
    \bibliographystyle{ieeenat_fullname}
    \bibliography{main}
}


\end{document}

%% file: 0_abstract.tex
\begin{abstract}
Recent studies in Large Vision-Language Models (LVLMs) have demonstrated impressive advancements in multimodal Out-of-Context (OOC) misinformation detection, discerning whether an authentic image is wrongly used in a claim. Despite their success, the textual evidence of authentic images retrieved from the inverse search is directly transmitted to LVLMs, leading to inaccurate or false information in the decision-making phase. To this end, we present E2LVLM, a novel evidence-enhanced large vision-language model by adapting textual evidence in two levels. First, motivated by the fact that textual evidence provided by external tools struggles to align with LVLMs inputs, we devise a reranking and rewriting strategy for generating coherent and contextually attuned content, thereby driving the aligned and effective behavior of LVLMs pertinent to authentic images. Second, to address the scarcity of news domain datasets with both judgment and explanation, we generate a novel OOC multimodal instruction-following dataset by prompting LVLMs with informative content to acquire plausible explanations. Further, we develop a multimodal instruction-tuning strategy with convincing explanations for beyond detection. This scheme contributes to E2LVLM for multimodal OOC misinformation detection and explanation. A multitude of experiments demonstrate that E2LVLM achieves superior performance than state-of-the-art methods, and also provides compelling rationales for judgments.\footnote{We will publicly release our codes.}
\end{abstract}

%% file: 1_intro.tex
\section{Introduction}
\label{sec:intro}

Recent years have witnessed the proliferation and advancement of Deepfake and social media platforms \cite{taigman2014deepface, liu2024forgery, qi2024sniffer, wang2024mmidr, shao2024detecting}, individuals freely present their speeches and opinions in form of image-text pairs. This inevitably accelerates the creation and ``viral'' spread of erroneous information, \ie, fake news, which intentionally deceives or misleads audiences in the real world. For example, during the COVID-19 pandemic~\cite{naeem2020covid}, falsified information conveyed by unaltered images and manipulated claims threats information security (IS) and societal order \cite{bohacek2024making, lin2024detecting}. Therefore, developing automated fact-checking specifically engineered for multimodal out-of-context (OOC) misinformation~\cite{abdelnabi2022open} has become a pressing necessity in the modern era.


\begin{figure}[t]
  \centering
   \includegraphics[width=1\linewidth]{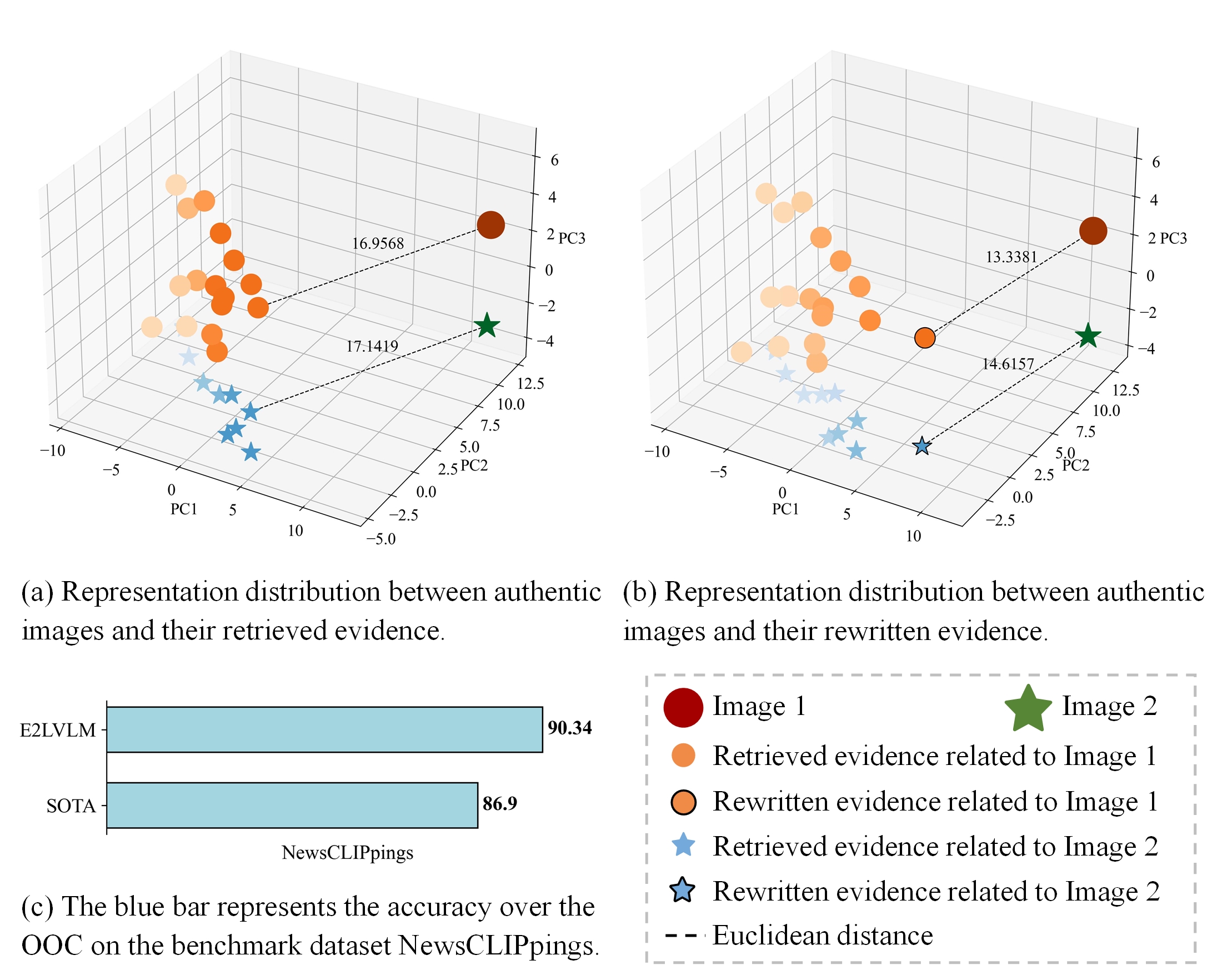}
   \caption{Subfigures (a) and (b) show the representation distributions between images and their retrieved/rewritten evidence. In addition, subfigure (c) provides a quantization comparison of E2LVLM over the OOC on NewsCLIPpings~\cite{luo2021newsclippings}.}
   \label{fig:1}
\end{figure}

Recent advancements \cite{yuan2023support, zhang2023ecenet, papadopoulos2023red} in the realm of OOC, the task of judging whether an authentic image is rightly or wrongly used in a claim, explore the use of external information retrieved by the custom search engine, such as Google API. Such methods achieve noteworthy improvements in performance. Unlike previous works without injecting external information \cite{aneja2023cosmos, mu2023self, dang2024overview}, where they solely focus on semantic similarity within image-claim pairs, a professional fact-checker commonly takes account of both internal and external information related to image-claim pairs for accurate decisions. A crucial challenge in the OOC task under the retrieved evidence lies in effectively leveraging the potential evidence to make the models identify inconsistencies regarding image-claim pairs.

Some efforts \cite{abdelnabi2022open, qi2024sniffer, tahmasebi2024multimodal} endeavor to introduce the retrieved evidence into attached classifiers, pre-trained models, and LVLMs, \eg, InstructBLIP~\cite{dai2023instructblip}, for the OOC task. This increases architectural complexity for better performance. However, a significant limitation of them is that the retrieved textual evidence doesn't match LVLMs inputs in the sense. As depicted in Figure~\ref{fig:1}, we have observed the following findings:
\begin{itemize}
    \item \textit{Not all textual evidence is effective. Relevant and irrelevant evidence are entangled, making the models challenging to fact verification.}
    \item \textit{A significant discrepancy exists between the retrieved textual evidence and rewritten content. LVLMs suffer from pieces of external information.}
\end{itemize}
This investigation indicates that existing retrieved textual evidence related to authentic images is relevant or irrelevant, which is not effective enough to empower the models with the discriminatory capacity over multimodal OOC misinformation. On the other hand, textual evidence reranking acquires the potentially available clue to support or refute the input image-claim pairs, while the representation of textual evidence is far from that of authentic images, as illustrated in subfigure (a) of \Cref{fig:1}. It is attributed that the retrieved textual evidence may include omit crucial details, redundant content, or fail to align with the input format of LVLMs~\cite{wu2024cotkr}. Therefore, exploring the reranking and rewriting of the retrieved textual evidence about authentic images is crucial to assist LVLMs for both judgment and explanation in the task of OOC.

In this paper, we present an innovative LVLMs-centric framework, dubbed E2LVLM, designed to deeply explore the use of textual evidence of authentic images for checking multimodal OOC misinformation in practical scenarios. Considering a trade-off between computation burden and information capacity~\cite{yin2023survey}, E2LVLM adopts the open-source LVLM such as Qwen2-VL~\cite{wang2024qwen2}, rather than closed-source LVLMs (\eg, GPT-4o~\cite{hurst2024gpt} and Gemini~\cite{team2023gemini}). E2LVLM depends on the vision and language understanding capabilities of Qwen2-VL to acquire informative content for aligning with LVLMs inputs. It first prompts Qwen2-VL to rerank textual evidence, and selects one most relevant textual evidence as external information. Then, E2LVLM rewrites the selected evidence to generate coherent and contextually attuned content, which eliminates the discrepancy between the retrieved textual evidence and model's logic. As shown in subfigure (b) of \Cref{fig:1}, the rewritten content based on the textual evidence reranking is closer to authentic images.

Moreover, to tackle the scarcity of the model’s explainability regarding multimodal OOC misinformation verification, E2LVLM incorporates the rewritten content into Qwen2-VL to generate explanation annotations for image-claim pairs, apart from ``Pristine'' and ``Falsified'' elements. Based on such supervised signal, E2LVLM attempts to extend one-stage multimodal instruction tuning on Qwen2-VL to the task of OOC. A large body of experiments shows that E2LVLM not only provides accurate decisions, but also yields compelling rationales for their judgments. As shown in subfigure (c) of \Cref{fig:1}, the proposed E2LVLM achieves an accuracy of 90.34\% over the OOC, and outperforms the state-of-the-art (SOTA) method by around 3.44\% accuracy on the benchmark dataset NewsCLIPpings~\cite{luo2021newsclippings}.

In summary, this paper has three-fold contributions as follows:
\begin{itemize}
    \item We investigate the discrepancy of representation distributions between images and their textual evidence, and underline the importance of the effective use of textual evidence related to images for better OOC detection.
    \item Based on this insight, we propose E2LVLM - a novel LVLMs-based framework, for multimodal OOC misinformation detection. Introducing the retrieved textual evidence into LVLMs reranks and rewrites them to acquire informative content for identifying the inconsistencies between image and claim.
    \item We further design a simple yet automated data generation pipeline, to construct a multimodal instruction-following dataset with both judgment and explanation.  Leveraging this dataset, we adopt a one-stage instruction tuning strategy to fine-tune E2LVLM for debunking multimodal OOC misinformation.
    \item We conduct extensive experiments evaluating E2LVLM on the public benchmark dataset NewsCLIPpings, and demonstrate its superiority in terms of both judgment and explanation. E2LVLM outperforms the current SOTA by around 3.44\% accuracy over the OOC.
    
\end{itemize}

%% file: 2_related.tex
\section{Related Work}
\label{sec:Related Work}

\begin{figure*}[t]
  \includegraphics[width=\linewidth]{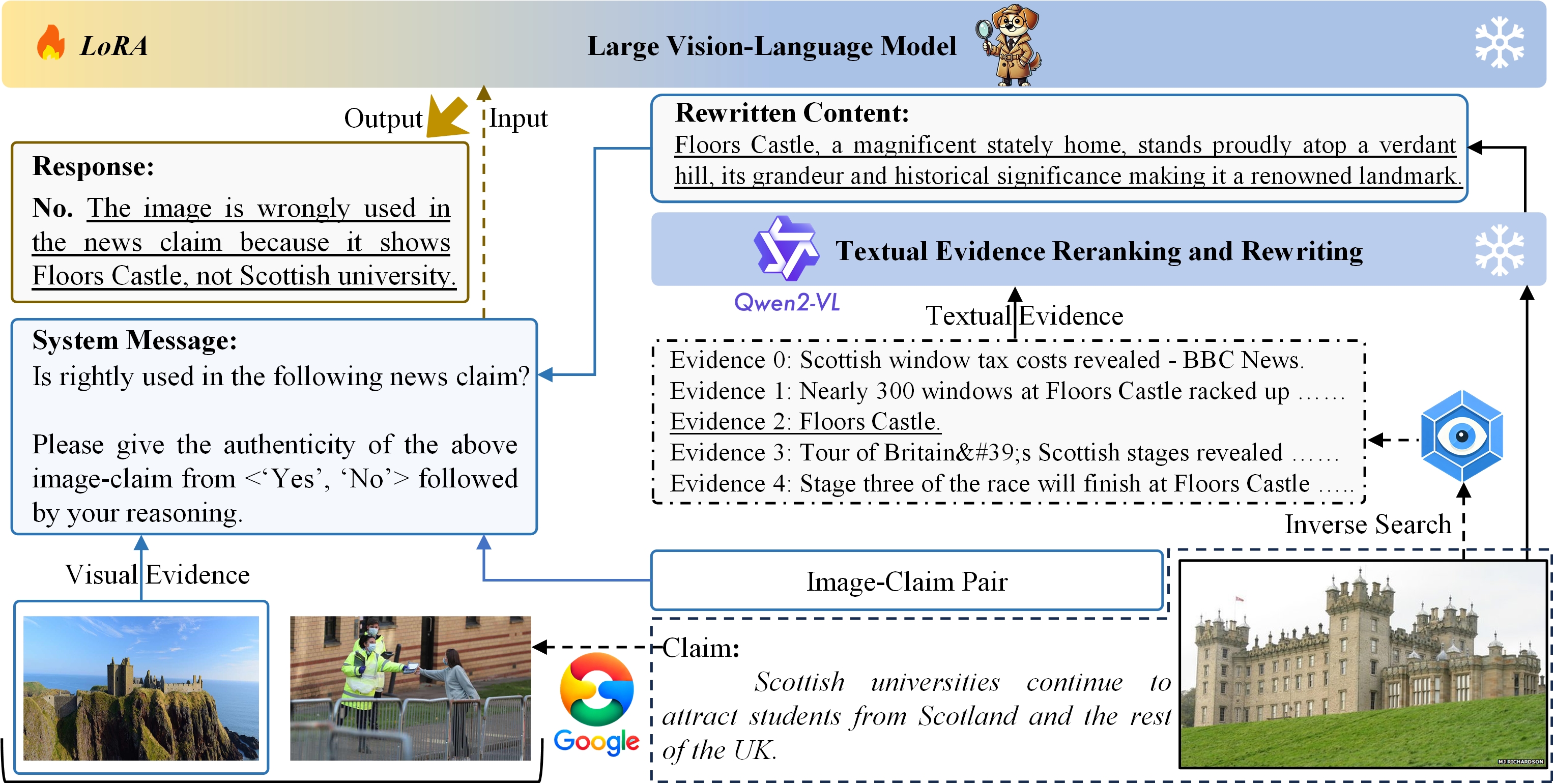}
  \caption{An overview of E2LVLM - the evidence-enhanced large vision-language model. Given an authentic image input together with a text claim, Google APIs are used to retrieve external evidence about the image-claim pair in an inverse search manner~\cite{abdelnabi2022open}. The image and its retrieved textual evidence are input to a Large Vision-Language model (Qwen2-VL) for evidence reranking. The top-1 textual evidence is further rewritten by the LVLM model. Apart from textual evidence, the retrieved visual evidence about the claim is reranked by cosine similarity, which achieves the top-1 visual evidence. Such content is input to E2LVLM together with the task-specific prompt for desired behaviors. Given this context, E2LVLM can provide its judgment and explanation for the authenticity of the image-claim pair.}
  \label{fig:2}
\end{figure*}

\subsection{Out-of-Context Misinformation Detection}
Due to the unprecedented growth of multimodal misinformation within social networks, many researches endeavor to address the issue of multimodal fact-checking, especially for multimodal OOC misinformation detection. Existing OOC methods can be roughly categorized into two orientations. Initial \cite{luo2021newsclippings, papadopoulos2023synthetic, gu2024learning} attempts to measure semantic similarity within image-claim pairs. Despite their impressive performance in the OOC task, there is a lack of exploration on the use of external information related to image-claim pairs. This results in the models with insufficient learning regarding logical or factual inconsistencies.

More recently, some methods \cite{zhang2023ecenet, wang2024mmidr, liu2024forgery} necessitate robust image-evidence modelling for debunking OOC misinformation. For example, Abdelnabi \etal~\cite{abdelnabi2022open} propose CNN, a pioneering work, to gather textual and visual evidence about image-claim pairs, for multimodal feature enhancement. Yuan \etal~\cite{yuan2023support} find the fact that the stance of external evidence induces a bias towards judgment. Further, Qi \etal~\cite{qi2024sniffer} firstly adopt the LVLM model with Vicuna-13B~\cite{vicuna2023} for addressing the OOC task in the era of generative artificial intelligence (AI) models. In contrast, our work explores the multimodal understanding capability of the open-source LVLM to rerank and rewrite textual evidence to make up support for debunking multimodal OOC misinformation.

\subsection{Instruction Tuning in the field of LVLMs}
Instruction tuning has emerged as a practical technology that is leveraged for fine-tuning LVLMs with instruction-following datasets, which presents promising performance improvements in different task-specific models \cite{zhang2023aligning, chen2024lion, jin2024llava}. More importantly, the successful adoption of knowledge distillation has paved the way for constructing instruction-following datasets \cite{yang2024gpt4tools, wang2024mmidr, kuckreja2024geochat}. The common practice generates instructional data from LVLMs in a query manner. Typically, LLaVA~\cite{dai2023instructblip} converts images into textual descriptions, and leverages ChatGPT (gpt-3.5-turbo)~\cite{ouyang2022training} for desired target behaviors. In the realm of OOC, Qi \etal~\cite{qi2024sniffer} adopts this idea, and employs language-only GPT-4~\cite{achiam2023gpt} to generate OOC instructional data as the supervised signal for subsequent training.

Differently, our work focuses on raw authentic images rather than translated textual descriptions. Furthermore, we incorporate the rewritten textual evidence with the image-claim pair, generating OOC multimodal instruction dataset. This may be more specialized for samples that include both judgments and explanations.

%% file: 3_methodology.tex
\section{Methodology}
\label{sec:Methodology}

In this section, we present the details of the Evidence-Enhanced Large Vision-Language Model (E2LVLM). The proposed E2LVLM endeavors to utilize textual evidence related to authentic images for debunking multimodal OOC misinformation. The framework is illustrated in \Cref{fig:2}.

\begin{figure*}
  \centering
      \includegraphics[width=\linewidth]{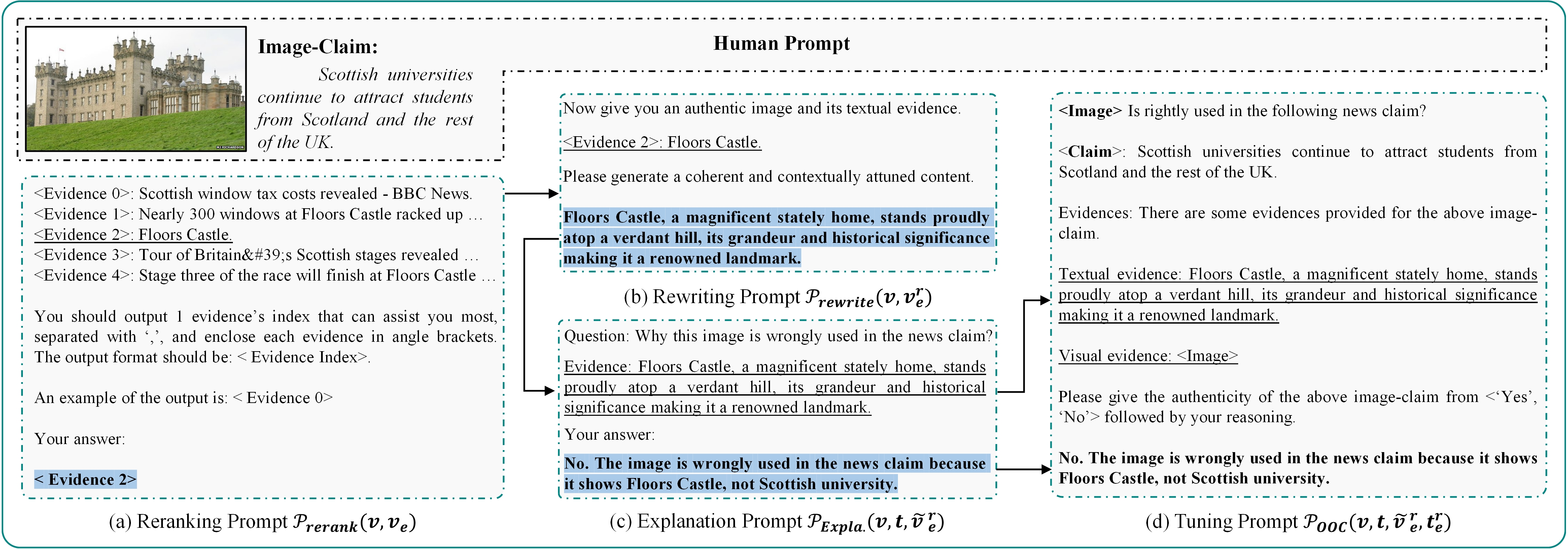}
    \caption{Prompts and their examples in E2LVLM. (a) Reranking prompt $\mathcal{P}_\mathrm{rerank}$ is to select one most relevant textual evidence related to the authentic image. (b) Rewriting prompt $\mathcal{P}_\mathrm{rewrite}$ is to achieve the coherent and contextually attuned content for alignment. (c) Explanation prompt $\mathcal{P}_\mathrm{Expla.}$ is to generate the compelling rationale to make up support for its assessment. (d) Tuning prompt $\mathcal{P}_\mathrm{OOC}$ is to extend the general-purpose LVLM to the task of multimodal out-of-context misinformation detection.}
    \label{fig:3}
\end{figure*}

\input{3.1_TDB}
\input{3.3_OOC_MID}
\input{3.4_E2FT}

%% file: 3.1_TDB.tex
\subsection{Task Definition and Background}

Given an authentic image $v$ and its claim $t$ from a dataset $\mathcal{D}$, where the image-claim pair $(v,t)\in\mathcal{D}$ achieves a multitude of evidence $(v_e, t_e)$ gathered by~\cite{abdelnabi2022open}, the task of OOC misinformation detection is to acquire a model $f_{ooc}(v,t,v_e, t_e)$ that makes a prediction $\hat{y}$ regarding the authenticity of the image-claim pair. Each pair is assigned with a semantic label $y\in\{0,1\}$, being either the ``Falsified'' (out-of-context) $(1)$ or ``Pristine'' (not out-of-context) $(0)$.

A typical approach to this issue is to train a classifier that provides a probability $p_{ooc}=p(\hat{y}|v,t,v_e, t_e)$. A more attractive solution taken by Qi \etal~\cite{qi2024sniffer} is to fine-tune the general-domain LVLM $g(v,t,v_e, t_e;\theta)$ on this task-specific dataset, thereby yielding the token $\mu$ of the corresponding semantic label. The symbol $\theta$ represents the learnable parameters of the LVLM model during training.

In the context of LVLMs, existing methods for discerning multimodal OOC misinformation typically default to the use of all the evidence $v_e$, $t_e$ of the image $v$ and claim $t$, but for image-evidence, the long distance between them inevitably prevents the model's discriminatory powers in the sense. Upon investigating the realm of OOC, we summarize the following findings:

\begin{enumerate}[label=\arabic*.]
    \item The fine-tuning model should not require access to all the evidence regarding image-claim pairs.
    \item The LVLMs-based model should require coherent and contextually attuned content, rather than pieces of external information.
    \item Similarly, the model should require both judgment and explanation, because this supervised signal contributes to the model's discriminatory powers.
\end{enumerate}

\subsection{Textual Evidence Reranking and Rewriting}

To rerank the gathered textual evidence of authentic images for acquiring relevant items, the most intuitive solution is to directly depend on the cosine similarity to calculate the relevance between the image $v$ and textual evidence $v_e=\{v^1_e, v^2_e,...,v^k_e\}$, \ie, $\mathrm{argsort}\ sim(\mathbf{z}_v, \mathbf{z}_{v^k_e})$, where $sim(\cdot)$ represents the cosine similarity, $\mathbf{z}_v\in\mathbb{R}^{1\times dim}$ and $\mathbf{z}_{v^k_e}\in\mathbb{R}^{1\times dim}$ denote the $dim$-dimensionality representations of the image and $k$-th evidence, respectively. However, this solution seriously lies in representation quality extracted by chosen backbone encoders (\eg, ViT B/32~\cite{radford2021learning} and ViT L/14~\cite{li2023blip}), and suffers from complex context. This is a sub-optimal option. To overcome this issue, we tend to multimodal understanding and reranking capabilities of LVLMs, and devise a textual evidence reranking strategy.

Textual evidence reranking refers to adopting the LVLM model Qwen2-VL~\cite{wang2024qwen2} to select one most relevant textual evidence $v^r_e$ related to the image. Due to the length extrapolation capability of Qwen2-VL, we incorporate the authentic image and its retrieved textual evidence as the part of the reranking prompt, denoted as $\mathcal{P}_\mathrm{rerank}(v, v_e)$. To finely control the output format, we attach a simple yet effective demonstration to the corresponding prompt. We present this prompt and its example in subfigure (a) of \Cref{fig:3}.

Further, in order to alleviate the discrepancy between the selected textual evidence $v^r_e$ and natural language-based LVLMs for alignment, we design a textual evidence rewriting strategy. Similar to the input prompt of the LVLM model for rerank, we leverage $(v, v^r_e)$ as part of the rewriting prompt, denoted as $\mathcal{P}_\mathrm{rewrite}(v, v^r_e)$, to guide the model for generating coherent and attuned content $\widetilde{v}^r_e$. Such prompt and its example are depicted in subfigure (b) of \Cref{fig:3}.

Additionally, we observe the fact that access to textual evidence of images is not invariably retrieved by search engines. In response to the lack of such textual evidence in the OOC realm, we directly employ the LVLM model to produce image captions as the rewritten content $\widetilde{v}^r_e$.

%% file: 3.3_OOC_MID.tex
\subsection{OOC Multimodal Instruction Dataset}

For the news domain and OOC detection task, a multimodal instruction dataset with both judgment and explanation is constructed to enhance the model's discriminatory powers. Although the previous method~\cite{qi2024sniffer} adopts InstructBLIP~\cite{dai2023instructblip} converting images into textual descriptions (as if it could visualize the image) and uses the close-source language-only GPT-4~\cite{achiam2023gpt} to capture inconsistencies between them, this destroys informative content related to authentic images. We summarize this issue stemming from a cross-modal semantic gap between images and the corresponding textual descriptions~\cite{jiang2024hallucination}. Therefore, we retain authentic images for claims in the OOC detection.

By leveraging Qwen2-VL~\cite{wang2024qwen2}, we adjust the model to follow an explanation prompt $\mathcal{P}_{Expla.}$, as shown in subfigure (c) of \Cref{fig:3}. Typically, we provide $(v,t, \widetilde{v}^r_e)$ as the part of this prompt that queries Qwen2-VL~\cite{wang2024qwen2} to perform desired behaviors, \ie, generating compelling explanations to support their judgments for the falsified information. The whole process can be defined as follows,
\begin{equation}\label{eq:vcg_5}
  \widetilde{\mathcal{D}} = \{(x,\widetilde{v}^r_e,{t}^r_e),s\mid x \sim \mathcal{D}, s \sim g(s \mid I(v,t,\widetilde{v}^r_e))\},
\end{equation}
where $I(v,t,\widetilde{v}^r_e)$ denotes an instantiated instruction by presenting an image-claim pair $(v,t)$ and the rewritten content $\widetilde{v}^r_e$. Besides, $g$ and $s$ refer to the LVLM model and its output, respectively. The sign $x$ is the sample $(v,t)$, and ${t}^r_e$ denotes the most relevant item related to the claim $t$. This item is achieved by cosine similarity, unless otherwise specified. As for NewsCLIPpings~\cite{luo2021newsclippings}, the constructed OOC multimodal instruction dataset $\widetilde{\mathcal{D}}$ consists of instructions with an equal number of falsified samples and pristine samples.

%% file: 3.4_E2FT.tex
\subsection{Evidence-enhanced Fine-tuning}

Regarding the identified challenge in the LVLM model Qwen2-VL~\cite{wang2024qwen2} when it comes to the OOC detection task, we suggest a one-stage multimodal instruction tuning solution extending the general-purpose LVLM, to the news domain for discerning multimodal OOC misinformation.

Our evidence-enhanced fine-tuning strategy is to provide both judgments and explanations for each image-claim pair. Simply put, we introduce both questions and candidate answers~\cite{shao2023prompting} into the tuning prompt $\mathcal{P}_\mathrm{OOC}(v,t,\widetilde{v}^r_e,{t}^r_e)$ that serves as informative inputs of the model. This makes the LVLM model unleash the potential knowledge behind itself~\cite{liu2024fka}, which explicitly analyzes the discrepancy between candidate answers for more accurate decisions. An illustration of the model response is shown in subfigure (d) of \Cref{fig:3}. For an image-claim pair with the ``Falsified'' label, we utilize the image $v$, claim $t$, rewritten textual evidence $\widetilde{v}^r_e$, and reranked visual evidence ${t}^r_e$ to format the input prompt of the LVLM model for response generation. The model provides the response likewise ``$<$Judgment$>$ $<$Explanation$>$'', where $<$Judgment$>$ indicates the symbol (\ie, ``No'') associated with candidate answers, and $<$Explanation$>$ is a coherent sentence serving as the compelling rationale for supporting its assessment.

In order to align the initial training way of the LVLM model, we employ LoRA technology~\cite{hu2022lora} and the next token prediction loss for assessing the model’s output error. Therefore, the learning objective of the proposed model E2LVLM over the OOC task, which can be described as,
\begin{equation}\label{eq:vcg_5}
    \mathcal{L}_{\mathrm{ooc}} = \sum_{i=1}^{N} -\log P(\epsilon_i \mid v_i,t_i,\widetilde{v_i}^r_e, {t_i}^r_e, \theta_{\mathrm{ooc}}),
\end{equation}
where, $\epsilon_i$ is considered as the generated tokens of both judgment and explanation corresponding to the formatted input $(v_i,t_i,\widetilde{v_i}^r_e,{t_i}^r_e)$, and $\theta_{\mathrm{ooc}}$ refers to the learnable parameters of E2LVLM during training. Besides, $N$ is the size of $\widetilde{\mathcal{D}}$.

%% file: 4_experiment.tex
\section{Experiment}
\label{sec:Experiment}

In accordance with existing multimodal out-of-context misinformation detection methods \cite{abdelnabi2022open, qi2024sniffer}, we conduct a multitude of experiments to demonstrate the effectiveness of the proposed E2LVLM on the public benchmark dataset NewsCLIPpings~\cite{luo2021newsclippings}. Typically, we focus on the six evaluation questions as follows:
\begin{enumerate}[label=\textbf{Q\arabic*:}]
    \item  How does E2LVLM perform in the task of multimodal OOC misinformation detection?
    \item  How does each procedure contribute to the E2LVLM's performance in detection?
    \item  Does E2LVLM provide accurate detections and compelling rationales for their judgments?
    \item  How impact are different sizes of the base LVLM on E2LVLM in detection?
    \item  Can E2LVLM be rapidly deployed at the stage of early detection?
    \item  How does E2LVLM perform on the other dataset?
\end{enumerate}

\subsection{Experimental Setup}

\textbf{Dataset.} We evaluate the efficacy of the proposed E2LVLM on the dataset NewsCLIPpings~\cite{luo2021newsclippings}. This dataset serves as the largest real-world multimodal misinformation detection benchmark. We follow the standard protocol \cite{abdelnabi2022open, yuan2023support, qi2024sniffer}, and report experimental results on the Merged/Balance subset. This subset consists of 71,072 training, 7,024 validation, and 7,264 testing, respectively.

\noindent \textbf{Compared Baselines.} To make a comprehensive performance evaluation, we compare the proposed E2LVLM with a series of representative methods. (1) A line of research focuses on attached classifiers trained from scratch, including SAFE~\cite{massarelli2019safe} and EANN~\cite{wang2018eann}. (2) Another line of research underlines the use of pre-trained models, containing VisualBERT~\cite{li2019visualbert}, CLIP~\cite{radford2021learning}, Neu-Sym detector~\cite{zhang2023detecting}, DT-Transformer~\cite{papadopoulos2023synthetic}, CCN~\cite{abdelnabi2022open}, SEN~\cite{yuan2023support}, and ECENet~\cite{zhang2023ecenet}. (3) Furthermore, in the era of LVLMs, SNIFFER~\cite{qi2024sniffer} is the first attempt to adopt a multimodal large language model for addressing the OOC task. More details of these methods can be provided in their official papers. 

\noindent \textbf{Evaluation Metrics.} We regard the multimodal OOC misinformation detection issue as a binary classification task. Following the standard process~\cite{abdelnabi2022open}, the accuracy over all samples (All), the accuracy over the OOC (Falsified), and not OOC (Pristine) are reported as the metrics during evaluation for a fair comparison.

\noindent \textbf{Implementation Details.} We choose Qwen2-VL-7B~\cite{wang2024qwen2} as the base LVLM, unless otherwise specified. We implement E2LVLM on PyTorch~\cite{paszke2019pytorch} version 2.3.1 with CUDA 12.2, and train it for 2 epochs on 4 NVIDIA GeForce RTX 3090 GPUs with 24G of memory. We adopt FlashAttention - 2~\cite{dao2024flashattention} for efficient training on the visual encoder and large language model. We use a batch size of 8 and a learning rate of $2\times 10^{-4}$. The models are optimized using AdamW~\cite{loshchilov2017decoupled} optimizer with a linear warmup and a cosine learning rate scheduler. Additionally, all experimental results are the average of three runs with no hyper-parameter searching.

\input{4.1_experiment_q1}
\input{4.2_experiment_q2}

\input{4.2_experiment_q3}
\input{4.2_experiment_q5}

%% file: 4.1_experiment_q1.tex
\subsection{RQ1: Comparison with SOTA Methods}

\Cref{tab:tab_1} presents the detailed comparison of E2LVLM with existing OOC methods on NewsCLIPpings. We use ``-'' for partial methods that do not release source codes or results. As shown in these results, we summarize the following findings: (1) As for the accuracy over ``All'', E2LVLM outperforms all methods by a large margin. Even for the state-of-the-art (SNIFFER), E2LVLM still outperforms it by around 1.5\% accuracy. (2) E2LVLM owns strong discriminatory powers, with an improvement of around 3.4\% over ``Falsified'', increasing the SOTA from 86.9\% to 90.3\%. (3) E2LVLM has a trade-off between ``Falsified'' and ``Pristine''. As reported by CCN~\cite{abdelnabi2022open}, a professional OOC misinformation detector should accurately identify both ``Falsified'' and ``Pristine'' samples. E2LVLM provides a closer distance between them, compared with the SOTA. This confirms that the improvement in the E2LVLM's performance on ``Falsified'' does not come at the cost of its performance on ``Pristine'', or vice versa. (4) With the increasing of architectural complexity, the performance of OOC misinformation detectors has been significantly enhanced, which is consistent with previous statements. In the context of LVLMs-based OOC methods, E2LVLM with Qwen2-VL-7B~\cite{wang2024qwen2} is superior to SNIFFER that depends on GPT-4~\cite{achiam2023gpt} and Vicuna-13B~\cite{vicuna2023}. These findings suggest the superiority of the proposed method E2LVLM on the OOC detection.

\begin{table}[t]
    \caption{Performance for accuracy compared to existing methods on NewsCLIPpings~\cite{luo2021newsclippings}. The best results are indicated in \textbf{bold}.}
      \centering
      \begin{adjustbox}{valign=c, max width=\columnwidth}
      \begin{tabular}{l|c|ccc}
        \toprule
        Methods & Venue & \textbf{All} & \textbf{Falsified} & \textbf{Pristine}  \\
        \hline
        SAFE~\cite{massarelli2019safe}  & PAKDD20 & 52.8 & 54.8 & 52.0  \\
        EANN~\cite{wang2018eann}  & SIGKDD18 & 58.1 & 61.8 & 56.2  \\
        \hline
        VisualBERT~\cite{li2019visualbert}  & arXiv19 & 58.6 & 38.9 & 78.4 \\
        CLIP~\cite{radford2021learning} & ICML21  & 66.0 & 64.3 & 67.7  \\
        Neu-Sym detector~\cite{zhang2023detecting} & arXiv23 & 68.2 & - & - \\
        DT-Transformer~\cite{papadopoulos2023synthetic} & MAD23 & 77.1 & 78.6 & 75.6  \\
        CCN~\cite{abdelnabi2022open} & CVPR22 & 84.7 & 84.8 & 84.5  \\
        SEN~\cite{yuan2023support} & EMNLP23 & 87.1 & 85.5 & 88.6  \\
        ECENet~\cite{zhang2023ecenet} & MM23 & 87.7 & - & - \\
        \hline
        SNIFFER~\cite{qi2024sniffer} & CVPR24 & 88.4 & 86.9 & \textbf{91.8}  \\
        \rowcolor{lightgreen} E2LVLM (\textit{Ours}) & & \textbf{89.9} & \textbf{90.3} & 89.4  \\
        \bottomrule
  \end{tabular}
  \end{adjustbox}
  \label{tab:tab_1}
\end{table}


%% file: 4.2_experiment_q2.tex
\subsection{RQ2: Ablation Study}

\begin{table}[t]
\caption{Ablation experiments for the E2LVLM. Evaluated on the NewsCLIPpings~\cite{luo2021newsclippings}. ``\#Evid.'' and ``\#Expla.'' represent the use of textual evidence and the supervised signal with explanations.}
\centering
\setlength{\tabcolsep}{1.5pt} 
\renewcommand{\arraystretch}{1.2} 
    \begin{adjustbox}{valign=c,max width=\columnwidth}
        \begin{tabular}{cccccc|ccc}
            \toprule
            Qwen2-VL & \#Evid. & Rerank & Rewrite& \#Expla. & Tuning & \textbf{All} & \textbf{Falsified} & \textbf{Pristine} \\
            \hline
            \cmark & \xmark & \xmark & \xmark & \xmark &\xmark & 69.1 & 54.4 & 83.9 \\
            \cmark & \cmark & \xmark & \xmark & \xmark &\xmark & 76.7 &	68.0 &	85.3 \\
            \hline
            \cmark & \xmark & \xmark & \xmark & \xmark &\cmark   & 78.9 & 73.8 & 84.2 \\
            \cmark & \cmark & \xmark & \xmark & \xmark &\cmark   & 83.0 & 77.1 & 88.9 \\
            \cmark & \cmark & \cmark & \xmark & \xmark &\cmark   & 87.7 & 86.5 & 88.7  \\
            \cmark & \cmark & \cmark & \cmark & \xmark &\cmark   & 88.5 & 87.7 & 89.1  \\
            \rowcolor{lightgreen} \cmark & \cmark & \cmark & \cmark &  \cmark &\cmark  & \textbf{89.9} & \textbf{90.3} & \textbf{89.4} \\
            \bottomrule
        \end{tabular}
    \end{adjustbox}
\label{tab:tab_2}
\end{table}

We conduct ablation experiments to analyze the effectiveness of primary procedures in E2LVLM. The results are shown in \Cref{tab:tab_2}, we can obtain the following findings:

\begin{itemize}
    \item We evaluate the impact of the use of textual evidence in zero-shot scenarios. As shown in the first two rows of \Cref{tab:tab_2}, the introduction of evidence achieves significant gains in performance, especially over ``Falsified''. Typically, the raw Qwen2-VL~\cite{wang2024qwen2} provides an accuracy of 69.1$\%$ over ``All'', which is higher than random guessing. It is attributed to the multimodal understanding capabilities of LVLMs. However, the base model struggles to discern falsified information, which necessitates robust methods to handle this challenge. Introducing textual evidence into the base model provides a performance improvement of 13.6$\%$ over the OOC. Besides, the accuracy on ``All'' reaches 76.7$\%$ (a 7.6$\%$ improvement in performance). This suggests the importance of the use of textual evidence in LVLMs for the OOC detection.
    \item In the second study, as shown in the remainder of \Cref{tab:tab_2}, each procedure contributes to the performance of E2LVLM. Typically, the task-specific fine-tuning technology extends LVLMs to the OOC, leading to promising results compared with zero-shot scenarios. Next, we conduct the textual evidence reranking strategy on E2LVLM, focusing on the salient item for debunking OOC misinformation. This results in an accuracy of 87.7$\%$ on ``All'', and shows a 9.4$\%$ improvement on ``Falsified'', while avoiding performance degradation on ``Pristine''. The reason is that such strategy appropriately eliminates the noise in the retrieved evidence. Further, we enforce the textual evidence rewriting strategy to generate coherent and contextually attuned content for better understanding, leading to an accuracy of 88.5$\%$ on ``All''. This suggests the importance of the reranking and rewriting of the retrieved textual evidence in E2LVLM for the OOC detection.
    \item Additionally, to understand the importance of the model’s explainability, we introduce the supervised signal with both judgment and explanation to the training phase, as shown in the last row of \Cref{tab:tab_2}. This results in an accuracy of 90.3$\%$ over ``Falsified'', higher than others. This is caused by the fact that such supervised signal enhances the model's discriminatory powers, thereby making the model provide accurate detections and attach them with compelling rationales for judgments. These results indicate that the supervised signal with both judgment and explanation contributes to the performance of E2LVLM.
\end{itemize}


%% file: 4.2_experiment_q3.tex
\subsection{RQ3: Detection and Beyond}


\begin{figure}[t]
  \centering
   \includegraphics[width=.9\linewidth]{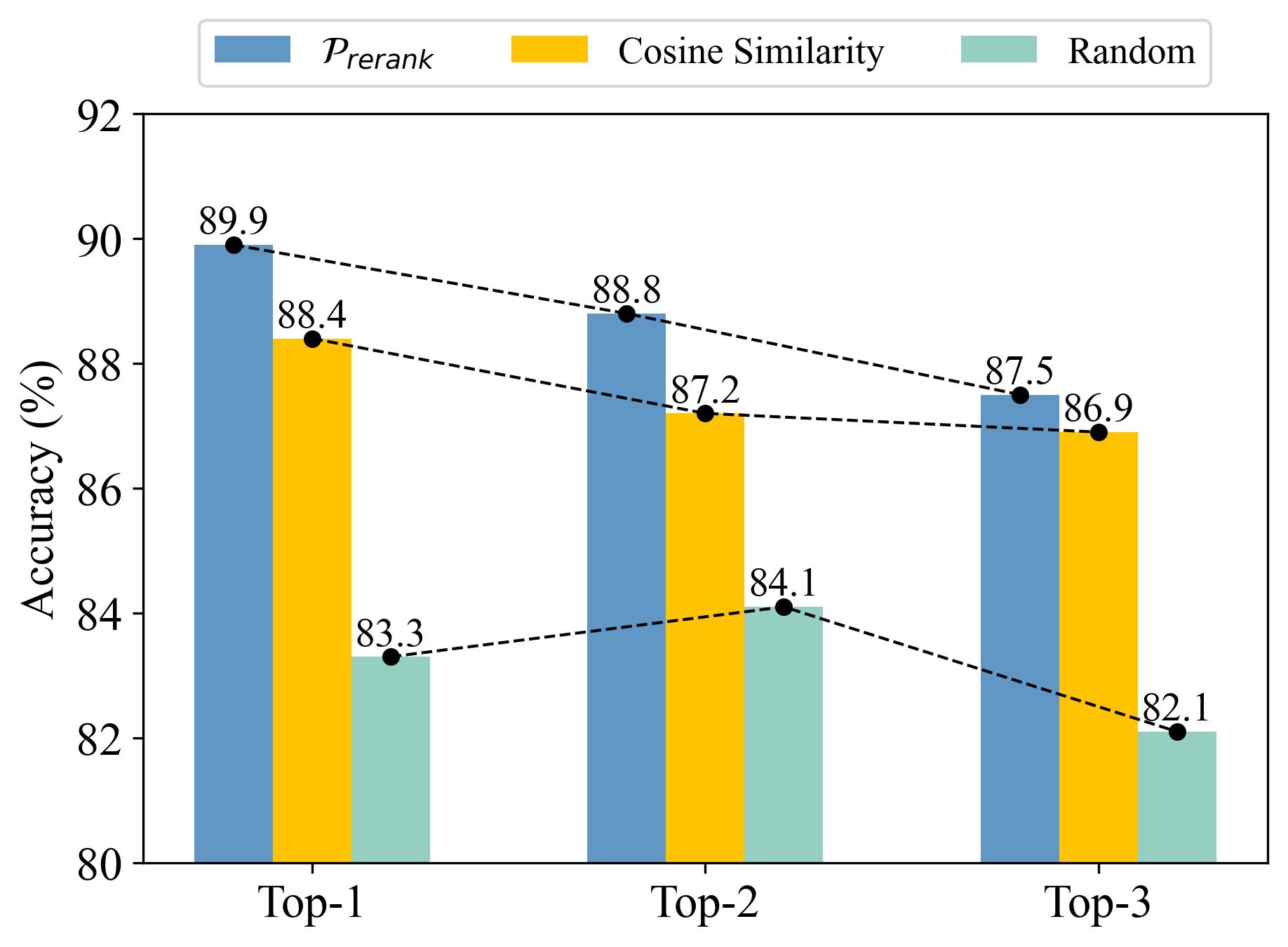}
   \caption{Comparison of E2LVLM with various reranking ways.}
   \label{fig:4}
\end{figure}


Considering authentic images in the OOC realm, we further evaluate the effectiveness of the textual evidence reranking and rewriting in E2LVLM for detection and beyond. In subsequent experiments, we uniformly adopt the accuracy over ``All'' samples for comparison, unless otherwise specified.

\noindent \textbf{Reranking Analysis.} Upon analysis experiments above, we have observed a significant performance of E2LVLM in the top-1 textual evidence. To understand the impact of this design, we start an investigation of E2LVLM's predictions, as shown in \Cref{fig:4}. We add other reranking ways for comparison, \ie, cosine similarity and random choice. For $k$, we set the range of it as $\{1,2,3\}$. As shown in these results, we can note that the change trends of accuracy are generally consistent across distinct ways, presenting degradations as the growth of textual evidence. This is caused by the fact that not all textual evidence is effective. This necessitates the textual evidence reranking to prevent the introduction of irrelevant items. Further, these results suggest that E2LVLM lies in the multimodal understanding and reranking capabilities of LVLMs, serving as a professional OOC misinformation detector that incorporates internal and external information for revealing misinformation.

\begin{figure}[t]
  \centering
   \includegraphics[width=1\linewidth]{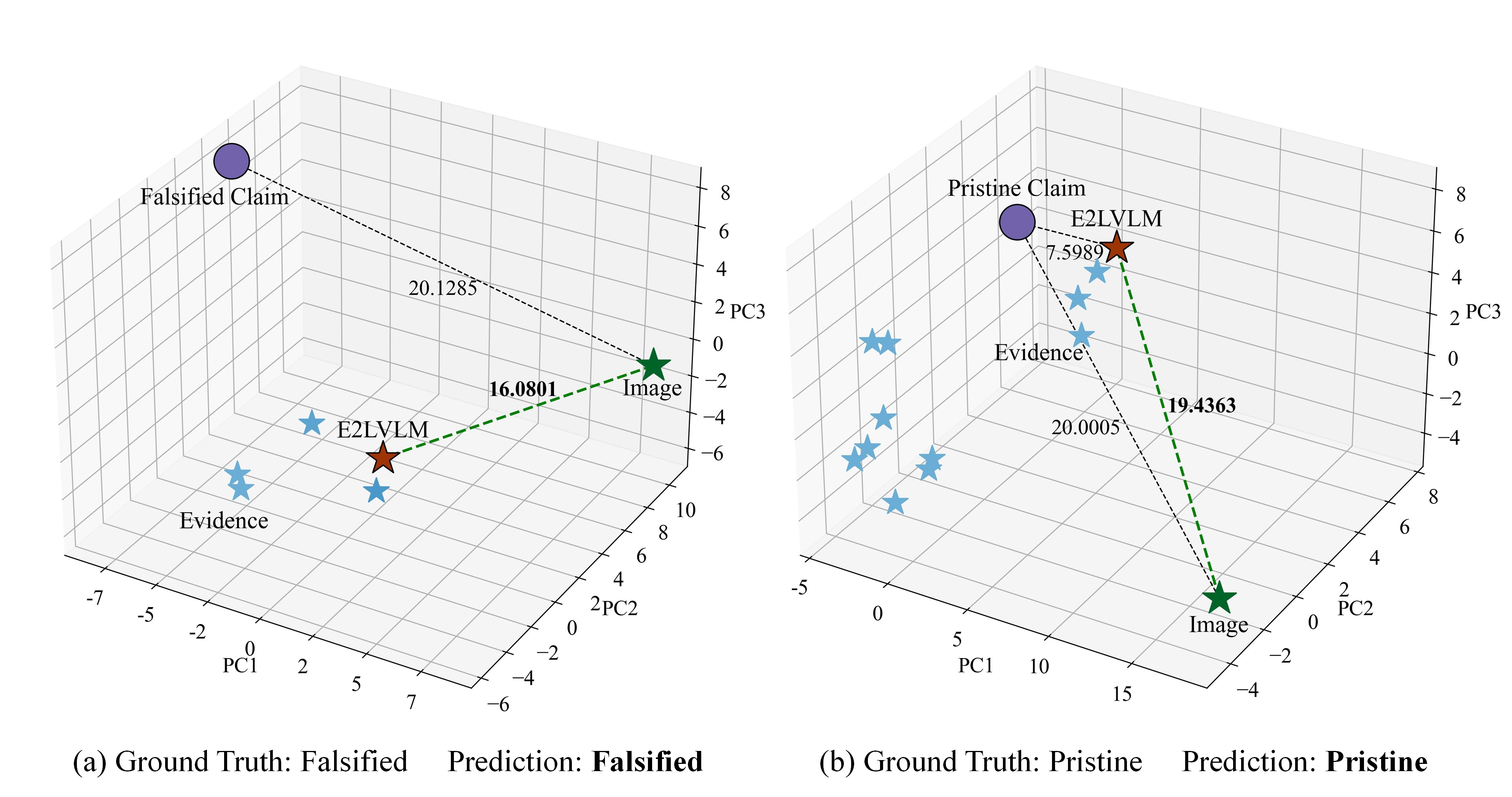}
   \caption{Visualization of various data distributions.}
   \label{fig:5}
\end{figure}

\noindent \textbf{Explainability Analysis.} To illustrate the E2LVLM's explainability, we conduct a qualitative analysis around data distributions to comprehend the decision-making phase, as shown in \Cref{fig:5}. From the results, the following findings are drawn: 1) As for an image-claim pair with ``Falsified'' in subfigure (a) of \Cref{fig:5}, the authentic image and its rewritten content are far from the falsified claim. Simultaneously, the rewritten content is closer to the authentic image, which supports it and refutes its claim. This makes the proposed E2LVLM provide an accurate decision. 2) As for an image-claim pair with ``Pristine'' in subfigure (b) of \Cref{fig:5}, the representation distributions of the authentic image, rewritten content, and claim are close together. This promotes E2LVLM to accurately discern this sample. These cases indicate that E2LVLM can effectively provide both judgment and explanation for debunking OOC misinformation.

\subsection{RQ4: Discussion of Different LVLMs}

To analyze the impact of distinct LVLMs on E2LVLM in the OOC detection, we conduct analysis experiments around Qwen2-VL family~\cite{wang2024qwen2} in \Cref{tab:tab_4} and \Cref{tab:tab_5}. As depicted in these results, we achieve the following observations:


(1) As shown in \Cref{tab:tab_4}, we employ different LVLMs for evidence-enhanced fine-tuning, apart from random guessing. The increasing of model's parameters provides better performance, which is consistent with previous research. Although the LVLM model Qwen2-VL-72B outperforms others (\eg, Qwen2-VL-7B) in a zero-shot scenario, it achieves a performance improvement of 0.7$\%$ with introducing around 10 times parameters. This ignores the compromise between computation burden and detection accuracy. Furthermore, the introduction of instruction tuning extends general-purpose LVLMs to the task of OOC, leading to significant improvements in performance (as shown in the fifth and last rows of this table).

(2) As depicted in \Cref{tab:tab_5}, we use different LVLMs at two stages, \ie, instructional data construction and model tuning. As for the former, we can observe that although LVLMs on a larger scale provide higher performance in detection accuracy, their improvements are finite (\eg, a performance improvement of 2.3$\%$ even in Qwen2-VL-7B). As for the latter, larger-scale LVLMs provide significant improvements in performance, leading to performance improvements of 6.5$\%$ and 8.5$\%$, respectively. This indicates the importance of the adopted LVLM in E2LVLM.

%% file: 4.2_experiment_q5.tex
\subsection{RQ5: Data Scaling Exploration}

\begin{table}[t]
\caption{A multimodal OOC misinformation detectors with distinct LVLMs. Evaluated on the NewsCLIPpings~\cite{luo2021newsclippings} dataset.}
\centering
\small
    \begin{adjustbox}{valign=c,max width=\columnwidth}
        \begin{tabular}{lc|c|c}
            \toprule
            Settings & Tuning & Parameters & \textbf{All} \\
            \hline
            Random & \xmark & -  & 50.3	\\
            Qwen2-VL-2B & \xmark & 1.5B  & 65.7 \\
            Qwen2-VL-7B & \xmark & 7.6B  &  79.1 \\
            Qwen2-VL-72B & \xmark & 72B  & 79.8 \\
            \hline
            Qwen2-VL-2B & \cmark & 1.5B  & 81.4	\\
            \rowcolor{lightgreen} Qwen2-VL-7B & \cmark & 7.6B  & \textbf{89.9} \\
            \bottomrule
        \end{tabular}
    \end{adjustbox}
\label{tab:tab_4}
\end{table}

\begin{table}[t]
\caption{Comparison on distinct LVLMs between data construction and model tuning. Evaluated on NewsCLIPpings~\cite{luo2021newsclippings}.}
\centering
\small
    \begin{adjustbox}{valign=c,max width=\columnwidth}
        \begin{tabular}{lcl|c}
        \toprule
         Data Construction &  \ding{223} & Model  Tuning & \textbf{All} \\
        \hline
        Qwen2-VL-2B & \ding{223} & Qwen2-VL-2B   & 81.1 \\
        Qwen2-VL-2B & \ding{223} & Qwen2-VL-7B   & 87.6 \\
        \hline
        Qwen2-VL-7B & \ding{223} & Qwen2-VL-2B   & 81.4 \\
        \rowcolor{lightgreen} Qwen2-VL-7B & \ding{223} & Qwen2-VL-7B   & \textbf{89.9} \\
        \bottomrule
        \end{tabular}
    \end{adjustbox}
\label{tab:tab_5}
\end{table}


To evaluate the feasibility of rapid deployment of E2LVLM at the stage of early detection, we randomly choose 10$\%$, 25$\%$, 50$\%$, and 75$\%$ samples for experiment analysis. As shown in \Cref{fig:6}, E2LVLM achieves remarkable performance on different proportions of training data. Typically, in a zero-shot scenario, E2LVLM has outperformed the previous method (\eg, DT-Transformer~\cite{papadopoulos2023synthetic}), suggesting the effectiveness of E2LVLM in the task of OOC. With the increasing of samples, the discriminatory powers of E2LVLM have a significant rising tendency. This demonstrates that the proposed E2LVLM can be rapidly deployed and ensure detection accuracy, even at the stage of early detection.

\subsection{RQ6: Robustness Analysis}



To further verify the robustness of E2LVLM towards real-world scenarios, we implement extended experiments on VERITE~\cite{papadopoulos2024verite}. This dataset serves as a novel real-world OOC misinformation, including 1,000 annotated samples stemming from fact-checking websites. Following~\cite{qi2024sniffer}, we also compare E2LVLM with RED-DOT~\cite{papadopoulos2023red} in \Cref{tab:tab_6}. From the results, we observe that E2LVLM shows the best performance on large-scale NewsCLIPpings~\cite{luo2021newsclippings} and more challenging VERITE~\cite{papadopoulos2024verite}. Typically, E2LVLM achieves a ``True vs OOC'' accuracy of 74.4$\%$, which outperforms RED-DOT by 0.5$\%$ accuracy, and the SOTA SNIFFER by 0.4$\%$ accuracy. The evaluations on these two multimodal OOC misinformation datasets exhibit the detection accuracy and robustness of E2LVLM in practical scenarios.

%% file: 5_conclusion.tex
\section{Conclusion}
\label{sec:Conclusion}

\begin{figure}[t]
  \centering
   \includegraphics[width=.9\linewidth]{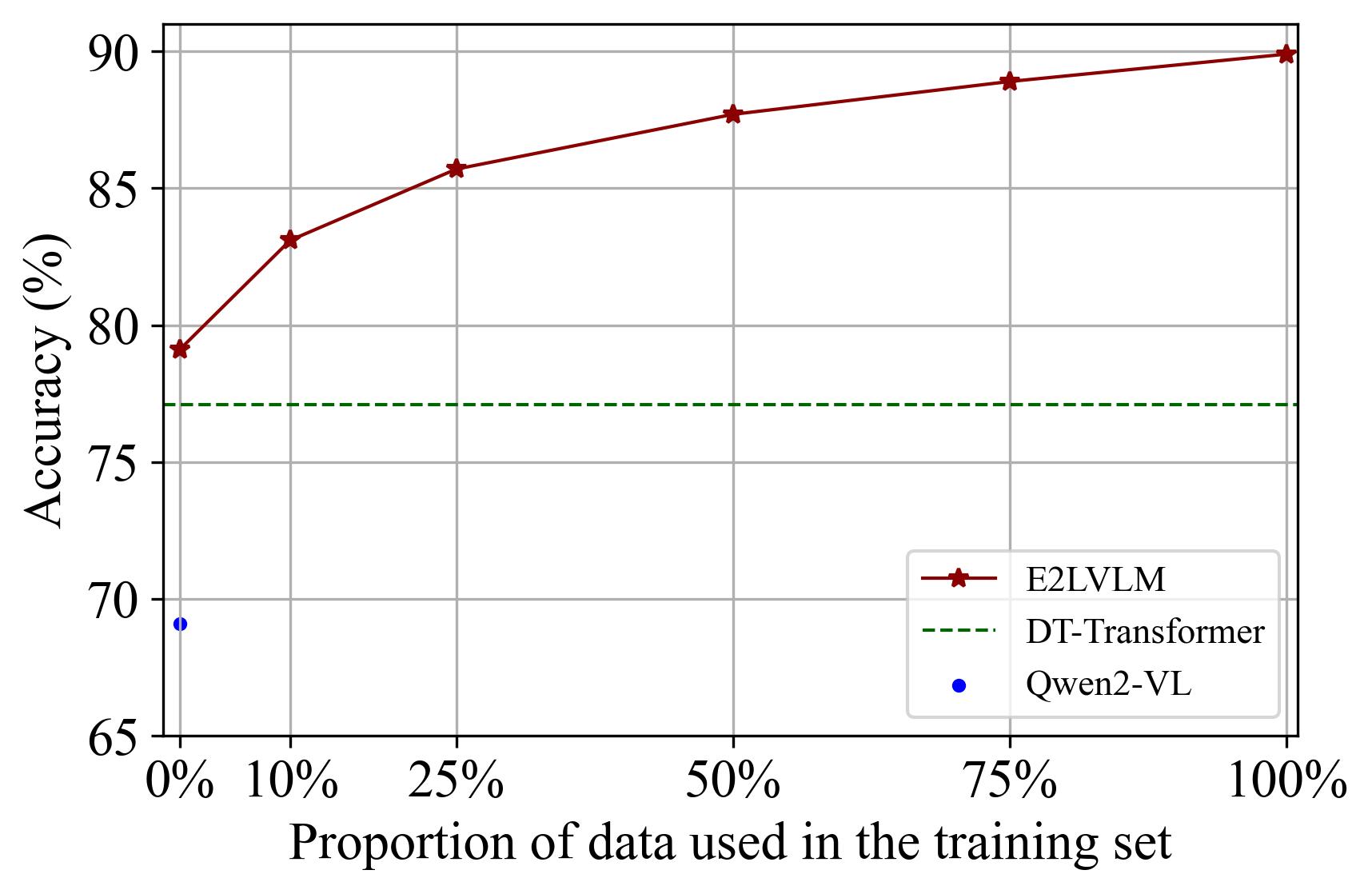}
   \caption{Performance of E2LVM on NewsCLIPpings~\cite{luo2021newsclippings} using different training data proportions.}
   \label{fig:6}
\end{figure}

\begin{table}[t]
\caption{Comparison of ``True vs OOC'' accuracy on the dataset VERITE~\cite{papadopoulos2024verite}. Trained on NewsCLIPpings~\cite{luo2021newsclippings} and evaluated on ``All''. The partial results are reported from SNIFFER~\cite{qi2024sniffer}.}
\centering
    \begin{adjustbox}{valign=c,max width=\columnwidth}
        \begin{tabular}{l|c|ll}
            \toprule
            Methods & Venue & \textbf{All} & \textbf{True vs OOC} \\
            \hline
            RED-DOT~\cite{papadopoulos2023red} & arXiv23 & 84.5 & 73.9 \\
            SNIFFER~\cite{qi2024sniffer} & CVPR24 & 88.4 ($\uparrow$ 3.9) & 74.0 ($\uparrow$ 0.1)  \\
            \rowcolor{lightgreen} E2LVLM (\textit{Ours}) &  & \textbf{89.9 ($\uparrow$ 5.4)} & \textbf{74.4 ($\uparrow$ 0.5)}  \\
            \bottomrule
        \end{tabular}
    \end{adjustbox}
\label{tab:tab_6}
\end{table}

Although advancements in LVLMs have provided promise in multimodal OOC misinformation detection, their performance is still limited due to the use of textual evidence regarding authentic images. Addressing this issue, we present E2LVLM, a novel evidence-enhanced LVLM for the OOC detection. E2LVLM not only provides accurate detections but also generates compelling rationales to support their judgments. As for retrieved textual evidence stemming from search engines, we rerank them to achieve one most relevant item, and then rewrite it to acquire coherent and contextually attuned content for LVLMs alignment. Furthermore, we construct the OOC multimodal instruction-following dataset with judgment and explanation. E2LVLM proposes a one-stage instruction tuning strategy to incorporate such a supervised signal into the LVLM, achieving impressive results between falsified and pristine samples. Experiments verify the superiority of E2LVLM over the OOC. In the context of LVLMs, we hope E2LVLM can provide potential schemes for making progress in the realm of OOC.
\clearpage